# Performance Analysis of Optimizers for Plant Disease Classification with Convolutional Neural Networks

Shreyas Rajesh Labhsetwar
*Department of Computer Engineering*
Fr. Conceicao Rodrigues Institute of Technology, Vashi
Navi Mumbai, India
shreyas.labhsetwar@fcrit.onmicrosoft.com

Soumya Haridas
*Department of Computer Engineering*
Fr. Conceicao Rodrigues Institute of Technology, Vashi
Navi Mumbai, India
soumya.haridas@fcrit.onmicrosoft.com

Riyali Panmand
*Department of Computer Engineering*
Fr. Conceicao Rodrigues Institute of Technology, Vashi
Navi Mumbai, India
riyali.panmand@fcrit.onmicrosoft.com

Rutuja Deshpande
*Department of Computer Engineering*
Fr. Conceicao Rodrigues Institute of Technology, Vashi
Navi Mumbai, India
rutuja.deshpande@fcrit.onmicrosoft.com

Piyush Arvind Kolte
*Department of Computer Engineering*
Fr. Conceicao Rodrigues Institute of Technology, Vashi
Navi Mumbai, India
piyush.kolte@fcrit.onmicrosoft.com

Sandhya Pati
*Department of Computer Engineering*
Fr. Conceicao Rodrigues Institute of Technology, Vashi
Navi Mumbai, India
sandhya.pati@fcrit.ac.in

*Abstract*— Crop failure owing to pests & diseases are inherent within Indian agriculture, leading to annual losses of 15-25% of productivity, resulting in a huge economic loss. This research analyzes the performance of various optimizers for predictive analysis of plant diseases with a deep learning approach. The research uses Convolutional Neural Networks for the classification of farm/plant leaf samples of 3 crops into 15 classes. The various optimizers used in this research include RMSprop, Adam, and AMSgrad. Optimizers' Performance is visualized by plotting the Training and Validation Accuracy and Loss curves, ROC curves, and Confusion Matrix. The best performance is achieved using Adam optimizer, with the maximum validation accuracy being 98%. This paper focuses on the research analysis proving that plant diseases can be predicted and pre-empted using deep learning methodology with the help of satellite, drone-based or mobile-based images that result in reducing crop failure and agricultural losses.

*Keywords*—Crop losses, Convolutional Neural Network, RMSprop, Adam, AMSgrad

## I. Introduction

India is a flourishing economic country, yet more than 60% of the inhabitants depend on farming or its associated products, either directly or indirectly, for their livelihood. Plant diseases due to pests lead to extreme loss of production and decline in the quality of the crop yield. Plant diseases are complicated, crop/region-specific, seasonal, epidemic/endemic, that need integrated approaches to manage the loss. Thanks to the extent of complexness and dimensions of land holdings, plant disease identification for preventive measures is difficult, including our inability to examine the pest/disease incidence and their life cycle with naked eyes. Due to the poor visibility of gadfly and illness occurrences, our ability to collect, store, integrate, and use the information for preventive/prescriptive measures has been a challenge.

Deep convolutional neural networks have achieved a significant improvement in classifying images. However, deep learning tasks require a vast amount of labeled data (qualitative as well as quantitative) for perfectly training the CNN models. The process of generating new homogeneous samples from the available dataset is called data augmentation, and it helps in enhancing its volume while also incorporating spatial invariance.

The primary objective of this research is to formulate an AI solution for plant disease prediction in large Indian farms where disease detection at the ultimate stages leads to crop produce failure, negative turnover, and in the worst cases, may lead to farmer suicides. A Predictive Deep Learning model will be an extremely quick, efficient, reliable, and cost-effective solution for plant disease detection. This research formulates a 28-layer Sequential CNN model to classify the plant images taken from satellites, drones or mobiles, into healthy and diseased categories. The dataset on which the model is trained consists of multiple high-resolution images belonging to the categories mentioned in TABLE 1. This research will facilitate the farmers to identify the percentage of the crop affected by pests and diseases, and in accordance with the extent of the disease, they can implement some of the solutions suggested by our software application to prevent the disease spread and thereby improve the crop yield.

## II. Related Work

Kun Guo et al. [1] analyzed the architecture style with hyper-parameters' improvement for convnets and observed few pragmatic principles for depth scaling on image classification tasks, which proved handy to resolve real problems and showcased experimentally that dropconnect layer is helpful for the regulation of wide-ranging neural networks. They demonstrated that training data noise can be controlled by the inculcation of slack variables in the cost function, thereby transforming the objective function to a new and more efficient version. Their experimental results showed that this new loss function (Hinge Loss) leads to improvement in the accuracy for classification tasks.

Yin xiaoj un et al.[2] determined the canopy spectral reflectance and DI for the detection of tomato bacterial spot disease and found the sensitivity band among the primitive spectral reflectance, first-order differential, second-order differential, and inverse logarithm spectral reflectance and validated the fact that the model of Second-order differential sensitive spectral reflectance is the best estimation model.

Priyadarshini Patil et al. [3] put forward an approach for predictive analysis of early & late blight diseases in potato farms with the exploitation of leaf images. They employed FCM clustering for the segmentation of disease-affected regions. Textural features from the diseased zones are extracted are utilized for classification. A thorough comparison based on the performance of various classifiers including Support Vector Machine, Random Forest, and Artificial Neural Networks is presented, with the highest accuracy of 92% achieved by the ANN algorithm. They emphasize the versatility of ANN while not having any restrictions on the input variables, and its ability to detect intricate hidden inter-variable dependencies.

Surampalli Ashok et al. [4] and Xuejian Liang et al. [7] proposed a CNN algorithm for extracting the features which match the test image intensities to the corresponding (true) classes and compare the same with the trained dataset images. This was achieved by optimizing & tuning various parameters of the leaves and thus reducing the classification error. The image classification technique was employed to compare the images and for their further classification. A DV-CNN model was used for HSI image classification consisting of small-sized labeled samples, which helps to increase the accuracy of classification in the neural network. In order to extract the features required for fusion, spectral and spatial data of leaves were used.

Jia Shijie et al. [5] discussed various data augmentation methodologies including Generative Adversarial Networks, Principal Component Analysis, Flipping, Shifting, Colour jittering, Noise Reduction, etc, to enhance the dataset quality for image classification task with CNNs, and showed its robustness in predicting the diseases by which crops are affected. The paper also highlights that the Wasserstein Generative Adversarial Network is an extremely efficient algorithm for image classification.

Sumathi Bhimavarapu et al. [6] proposed multiple algorithms for each scenario of the Transfer Learning Convolutional Neural Network for proper training and validation on different subsets of the dataset. It was observed that on different sets of input data, AlexNet and GoogLeNet demonstrated a very high tendency of overfitting for plant disease classification tasks. Here, problems were taken from varying sets of conditions so that model is optimal and scalable.

Zijun Zhang [8] developed a variation of Adam optimizer to get rid of the generalization gap. The methodology titled, 'normalized direction-preserving Adam (NDAdam)', allowed more precise control of the direction and step size for the updation of weights, resulting in significant improvement in the performance for generalization. Regularizing the softmax logits also contributed to the same. He introduced the NDAdam algorithm, a modified variant of the adam optimizer for training deep neural networks. NDAdam is implemented to maintain the direction of the gradient for each weight vector, and introduce the regularization effect of L2 weight decay in a more accurate and proper manner.

Priyanka Sharma et al. [9] proposed Artificial Neural Networks to be used over supervised learning methodologies so that the accuracy for late blight in potatoes can be improved further. They employed several different datasets for experiments. The principle aim was analyzing the robustness of various activation functions and it was observed that maximum accuracy was obtained in the case of the sigmoid function. This helped in suggesting that a hybrid combination of ANN (deep learning) with traditional machine learning algorithms may result in more efficient models for future prediction tasks based upon historical data and thus would help in saving the crops from getting infected. Moreover, they concluded that the prediction accuracy of ANN models is directly proportional to the size of the training and validation datasets.

III. PROPOSED WORK

This research aims to design and implement an AI-based Deep CNN model capable of performing predictive analysis of plant diseases based on satellite, drone-based, or mobile-based images. The model is trained on an open-source Plant Village Dataset. The CNN model is a custom implemented 28-layer Sequential Model with 15-way softmax activation in the last layer. Also, the research compares the performance of various optimizers, and suggest thereafter the most befitting ones for plant disease detection and classification.

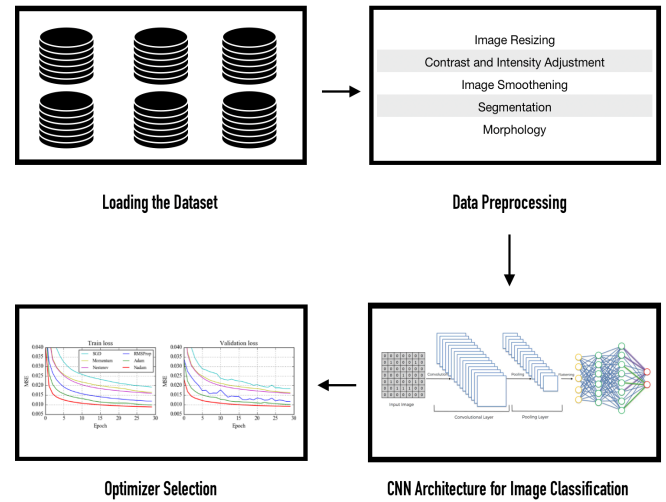

Fig. 1. Proposed Work Summary

The training is performed on an experiment based custom formulated CNN model comprising of 28 layers consisting of a hierarchy of Convolution, Max Pooling, Dropout, and Batch Normalization layers. The model is a high-width and high-depth CNN architecture including a maximum of 1024 neurons/layer. The stride size for each convolution layer is set to (3 X 3), the same being the pool size for MaxPooling layers. The performance of three different optimizers, namely, RMSprop, Adam, and AMSgrad are analyzed and compared. The performance is visualized by using the Confusion Matrix, Training and Validation Accuracy and Loss curves, and ROC curves. To finding the best hyperparameters for CNN, hyperparameter tuning is performed using Grid Search Cross-Validation Methodology [10].

*A. Dataset*

The dataset used for this research is an open-source Plant Village dataset consisting of 54303 healthy and diseased leaf images split into 38 categories based on the species and disease. After careful analysis of the dataset, the research is advanced focusing on 15 plant categories (classes) as mentioned in TABLE 1. The images are of high resolution and in RGB format. Since some classes have a low number of images, image augmentation techniques are employed to ensure that the count of images available per class is exactly 2000.

TABLE 1. Dataset Overview

| Class No. | Class Label | No. of Images |
|---|---|---|
| 1 | Tomato Late Blight | 1909 |
| 2 | Pepper Bell Healthy | 1478 |
| 3 | Tomato Septoria Leaf Spot | 1771 |
| 4 | Potato Late Blight | 1000 |
| 5 | Potato Early Blight | 1000 |
| 6 | Potato Healthy | 1520 |
| 7 | Tomato Healthy | 1591 |
| 8 | Tomato Leaf Mold | 952 |
| 9 | Tomato Yellow Leaf Curl Virus | 3209 |
| 10 | Tomato Bacterial Spot | 2127 |
| 11 | Tomato Mosaic Virus | 1730 |
| 12 | Tomato Target Spot | 1404 |
| 13 | Tomato Early Blight | 1000 |
| 14 | Pepper Bell Bacterial Spot | 997 |
| 15 | Tomato Spider Mites | 1676 |

*B. Preprocessing*

Agricultural Images taken using drones, satellites, or other means are often contaminated with noise. The noise may be a result of multiple factors including corpuscular nature of light, hardware noise due to mechanical issues in cameras, as well as natural factors including humans, animals in the captured images [11] which may negatively impact the results of experiments and contribute to false positives and negatives (FPR, FNR).

Thus, all images comprising the dataset are first preprocessed to suppress any unwanted distortions and also enhance the critical image features. The various types of preprocessing techniques employed as part of this study are:

1. Image Resizing
2. Contrast and Intensity Adjustment
3. Removal of Noise (Image Smoothening)
4. Segmentation
5. Morphology

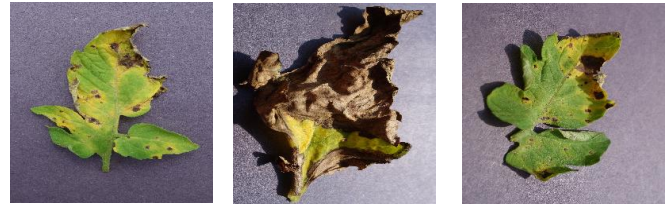

Fig. 2. Tomato Septoria Leaf Spot Dataset Images

All the images are rescaled to size (256 X 256) for dataset uniformity, reduction of computational complexity and help in feature extraction [12].

The prime objective of the research is to extract the diseased portions of leaves from all the images. For this purpose, the images are first made noise-free by image smoothening using Gaussian Blur Technique [13,14,15], and thereafter, all the RGB images are converted to the HSV colour space.

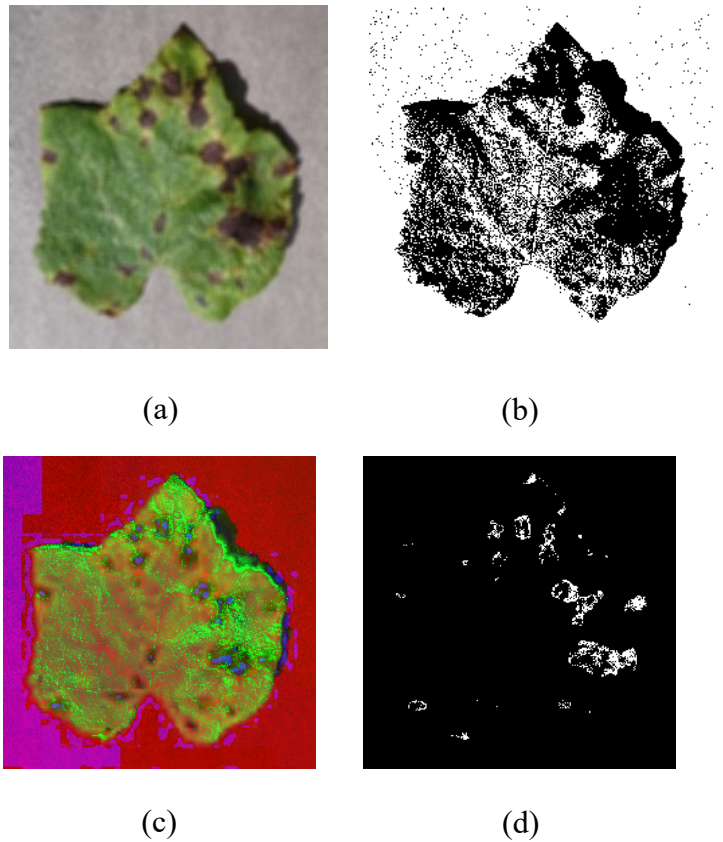

(a)     (b)

(c)     (d)

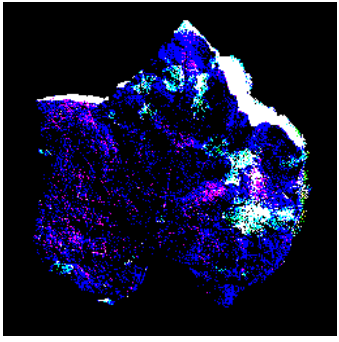

(e)

Fig. 3. (a) Gaussian Blur (b) Grayscale Image Thresholding (c) RGB to HSV Image (d) HSV Image Thresholding (e) RGB Image Thresholding

C. *Convolutional Neural Network*

- The proposed research involves the use of CNN for the classification of images into the respective classes.

- CNN is a very strong Data Mining Algorithm that employs deep learning methodology for image classification. The complexity of a Neural Network Algorithm depends on the task at hand.

- A CNN typically comprises a hierarchy of Convolution and Max Pooling Layers, and a Flatten Layer before the dense ANN.

- The Convolution Layer applies multiple feature detectors to the input image in order to generate corresponding Feature Maps [16].

- There can be multiple such feature detectors or filters applied in the Convolution Layer such as sharpen, blur, edge enhance, edge detect, emboss, etc.

- To this Convolution Layer, Rectified Linear Unit (ReLu) activation functions are applied to increase Non-Linearity in the CNN Model, thereby reducing overfitting.

- After this comes the Max Pooling Layer which outputs a set of Pooled Feature Maps that make up the Pooled Layer.

- Max Pooling is required to introduce spatial invariance in the CNN model. Thus, the classification will remain independent of spatial orientation factors such as the angle at which the image is taken, the angle at which the leaf is present, the angle and amount of sunlight falling, etc.

- The CNN model used in the proposed research consists of 28 layers as shown in Fig. 4.

- Flatten function is used to convert the Pooled Feature Maps to NumPy vectors which are fed as input to the ANN.

- For Hyperparameter Tuning, Grid-Search Cross-Validation is used to figure out the best set of hyperparameters that contribute to maximum accuracy [17,18].

| Layer (type) | Output Shape |
|---|---|
| Conv2D | (None, 256, 256, 32) |
| Activation | (None, 256, 256, 32) |
| Batch_Normalization_13 | (None, 256, 256, 32) |
| MaxPooling2 | (None, 85, 85, 32) |
| Dropout | (None, 85, 85, 32) |
| Conv2D | (None, 85, 85, 64) |
| Activation | (None, 85, 85, 64) |
| Batch Normalization 14 | (None, 85, 85, 64) |
| Conv2D | (None, 85, 85, 64) |
| Activation | (None, 85, 85, 64) |
| Batch Normalization 15 | (None, 85, 85, 64) |
| MaxPooling2 | (None, 42, 42, 64) |
| Dropout | (None, 42, 42, 64) |
| Conv2D | (None, 42, 42, 128) |
| Activation | (None, 42, 42, 128) |
| Batch Normalization 16 | (None, 42, 42, 128) |
| Conv2D | (None, 42, 42, 128) |
| Activation | (None, 42, 42, 128) |
| Batch Normalization 17 | (None, 42, 42, 128) |
| MaxPooling2 | (None, 21, 21, 128) |
| Dropout | (None, 21, 21, 128) |
| Flatten | (None, 56448) |
| Dense | (None, 1024) |
| Activation | (None, 1024) |
| Batch Normalization 18 | (None, 1024) |
| Dropout | (None, 1024) |
| Dense | (None, 15) |
| Activation | (None, 15) |

Fig. 4. CNN Architecture

D. *Optimizers*

Optimizers are algorithms which help compute the errors upon forward propagations and thus help in adjusting the features of a neural network, such as its weights and learning rate, thereby reducing the loss [19].

- RMSprop

  RMSprop Optimizer employs a dynamic learning rate which results in superior performance as compared to Adagrad optimizer by taking an exponential moving average of gradients rather than considering the cumulative sum of squared gradients (Adagrad) [20].

$$\omega(t+1) = \omega(t) - \frac{\alpha}{\sqrt{\lambda(t) + \varepsilon}} * \frac{\partial \Delta}{\partial \omega(t)}$$

Where,
$$\lambda(t) = \beta\lambda(t-1) + (1-\beta)[\frac{\partial\Delta}{\partial\omega(t)}]^2$$

$\lambda$ is initialised to 0,
$\beta = 0.95$,
$\varepsilon$ is the Regularization Term

- Adam

Adam Optimizer introduces the property of momentum in the RMSprop Optimizer. It regulates the gradient component with respect to the dynamic mean of slopes ($\mu$), and the learning rate component with respect to $\sqrt{\lambda}$, which represents the exponential dynamic mean of squared gradients (similar to RMSprop) [21,22].

$$\omega(t+1) = \omega(t) - \frac{\alpha}{\sqrt{\hat{\lambda}(t)} + \varepsilon} * \widehat{\mu(t)}$$

Where,
$$\widehat{\mu(t)} = \frac{\mu(t)}{1 - \beta1(t)}$$
$$\widehat{\lambda(t)} = \frac{\lambda(t)}{1 - \beta2(t)}$$
$$\mu(t) = \beta1\mu1(t-1) + (1-\beta1)\frac{\partial\Delta}{\partial\omega(t)}$$
$$\lambda(t) = \beta2\lambda(t-1) - (\beta2 - 1)[\frac{\partial\Delta}{\partial\omega(t)}]^2$$

$\mu$ and $\lambda$ are initialised to 0,
$\alpha$ is the Learning Rate,
$\beta1 = 0.9$ (Keras),
$\beta2 = 0.999$ (Keras),
$\varepsilon$ is the Regularization Term

- AMSgrad

AMSgrad Optimizer is a variant of Adam Optimizer which uses the dynamic learning rate property of Adam Optimizer and modifies it to ensure that the current $\lambda$ is always larger than the previous $\lambda$ (ever-increasing) [23].

$$\omega(t+1) = \omega(t) - \frac{\alpha}{\sqrt{\hat{\lambda}(t)} + \varepsilon} * \mu(t)$$

Where,
$$\widehat{\lambda(t)} = \max(\widehat{\lambda(t-1)}, \widehat{\lambda(t)})$$
$$\mu(t) = \beta1\mu1(t-1) + (1-\beta1)\frac{\partial\Delta}{\partial\omega(t)}$$
$$\lambda(t) = \beta2\lambda(t-1) + (1-\beta2)[\frac{\partial\Delta}{\partial\omega(t)}]^2$$

$\mu$ and $\lambda$ are initialised to 0,
$\alpha$ is the Learning Rate,
$\beta1 = 0.9$,
$\beta2 = 0.999$,
$\varepsilon$ is the Regularization Term

## IV. RESULTS

The research is performed on the aforementioned optimizers and their performance is visualized by plotting the Confusion Matrix and ROC Curve for each optimizer.

### A. RMSprop Optimizer

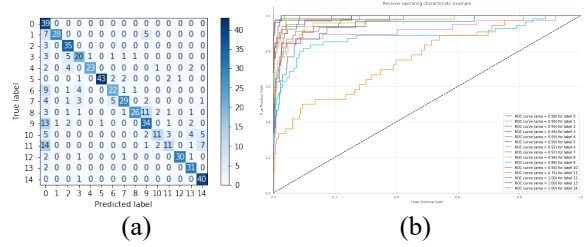

(a)      (b)

Fig. 5. (a) Confusion Matrix (b) ROC curve for RMSprop Optimizer

From Fig. 5. (a), it is observable that RMSprop performs perfect classification of almost all classes except it misclassifies Tomato Target Spot as Tomato Late Blight.

### B. Adam Optimizer

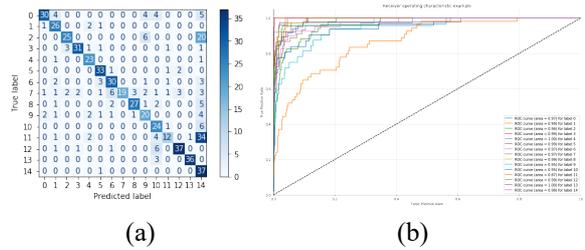

(a)      (b)

Fig. 6. (a) Confusion Matrix (b) ROC curve for Adam Optimizer

From Fig. 6. (a), it is evident that Adam performs perfect classification of almost all classes except it misclassifies Tomato Target Spot as Potato Late Blight.

### C. AMSgrad Optimizer

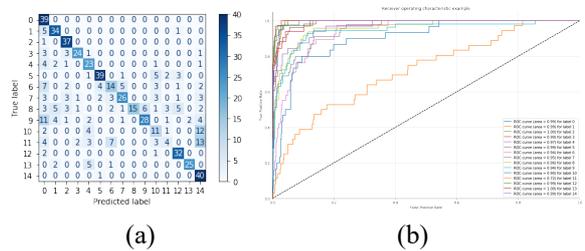

(a)      (b)

Fig. 7 (a) Confusion Matrix (b) ROC curve for AMSgrad Optimizer

From Fig. 7 (a), it is evident that AMSgrad performs perfect classification of most classes except it misclassifies Tomato Mosaic Virus as Potato Late Blight, Tomato Target Spot as Potato Late Blight.

## V. Conclusion

The proposed 28-layer Sequential CNN model is experimented with different optimizers and its performance is evaluated with multiple performance graphs including Confusion Matrix and ROC curve. The best performance is achieved by Adam optimizer, with the maximum validation accuracy being 98%. It is closely followed by RMSprop optimizer, with a 95% validation accuracy. This research thus analyses the performance of various optimizers for plant disease classification task and proves that plant diseases can be predicted and pre-empted using deep learning methodology on satellite, drone-based, or mobile-based images, and thus, reduce crop failure and pre-empt agricultural losses.